\documentclass[sigconf]{acmart}

\usepackage[T1]{fontenc}
\usepackage{aecompl}

\usepackage{amsmath, amsthm, amssymb}
\usepackage{bbm}
\usepackage{bbding}
\usepackage{threeparttable}
\usepackage{makecell}
\usepackage{multirow}
\usepackage{graphicx}
\usepackage{float}
\usepackage{enumitem} 
\usepackage{enumerate}
\usepackage{pifont}
\usepackage{caption}
\usepackage{balance}
\usepackage{setspace}
\usepackage{graphicx}
\usepackage[T1]{fontenc}
\usepackage{color}

\usepackage{aecompl}

\usepackage{url}

\usepackage{subcaption}
\usepackage{colortbl}
\usepackage{diagbox}

\usepackage[capitalize]{cleveref}
\crefname{section}{Sec.}{Secs.}
\Crefname{section}{Section}{Sections}
\Crefname{table}{Table}{Tables}
\crefname{table}{Tab.}{Tabs.}

\usepackage{xcolor}
\usepackage{hyperref}
\hypersetup{colorlinks,
      linkcolor=ACMPurple,
      citecolor=ACMPurple,
      urlcolor=ACMDarkBlue,
   filecolor=ACMDarkBlue
}
\usepackage{stmaryrd}
\usepackage{enumitem}
\usepackage{colortbl}
\def\ie{\textit{i.e.}}
\def\eg{\textit{e.g.}}

\begin{document}

\title{AdvCLIP: Downstream-agnostic Adversarial Examples in Multimodal Contrastive Learning}

\author{Ziqi Zhou}
\email{zhouziqi@hust.edu.cn}
\affiliation{%
  \institution{School of Cyber Science and Engineering, Huazhong University of Science and Technology} 
  \country{}
}

\authornote{National Engineering Research Center for Big Data Technology and System}
\authornote{Services Computing Technology and System Lab}
\authornote{Hubei Key Laboratory of Distributed System Security}
\authornote{Hubei Engineering Research Center on Big Data Security}


\author{Shengshan Hu}
\email{hushengshan@hust.edu.cn}
\affiliation{%
  \institution{School of Cyber Science and Engineering, Huazhong University of Science and Technology} 
\country{}
}
\authornotemark[1]
\authornotemark[2]
\authornotemark[3]
\authornotemark[4]

\author{Minghui Li}
\email{minghuili@hust.edu.cn}
\affiliation{%
  \institution{School of Software Engineering, Huazhong University of Science and Technology} 
\country{}
}

\author{Hangtao Zhang}
\email{zhanghangtao7@163.com}
\affiliation{%
  \institution{School of Cyber Science and Engineering, Huazhong University of Science and Technology} 
\country{}
}

\author{Yechao Zhang}
\email{ycz@hust.edu.cn}
\affiliation{%
  \institution{School of Cyber Science and Engineering, Huazhong University of Science and Technology} 
\country{}
}
\authornotemark[1]
\authornotemark[2]
\authornotemark[3]
\authornotemark[4]

\author{Hai Jin}
\email{hjin@hust.edu.cn}
\affiliation{%
  \institution{School of Computer Science and Technology, Huazhong University of Science and Technology} 
\country{}
}
\authornotemark[1]
\authornotemark[2]
\authornote{Cluster and Grid Computing Lab}

\renewcommand{\shortauthors}{Ziqi Zhou et al.}

\begin{abstract}
Multimodal contrastive learning aims to train a general-purpose feature extractor, such as CLIP, on vast amounts of raw, unlabeled paired image-text data.
This can greatly benefit various complex downstream tasks, including cross-modal image-text retrieval and image classification. Despite its promising prospect, the security issue of cross-modal pre-trained encoder has not been fully explored yet, 
especially when the pre-trained encoder is publicly available for commercial use.

In this work, we propose AdvCLIP, the first attack framework for generating downstream-agnostic adversarial examples based on cross-modal pre-trained encoders. 
AdvCLIP aims to construct a universal adversarial patch for a set of natural images that can fool all the downstream tasks inheriting the victim cross-modal pre-trained encoder.
To address the challenges of heterogeneity between different modalities and unknown downstream tasks, 
we first build a topological graph structure to capture the relevant positions between target samples and their neighbors. 
Then, we design a topology-deviation based generative adversarial network to generate a universal adversarial patch. By adding the patch to images, we minimize their embeddings similarity to different modality and perturb the sample distribution in the feature space, achieving unviersal non-targeted attacks.
Our results demonstrate the excellent attack performance of AdvCLIP on two types of downstream tasks across eight datasets. We also tailor three popular defenses to mitigate AdvCLIP, highlighting the need for new defense mechanisms to defend cross-modal pre-trained encoders.
Our codes are available at: \url{https://github.com/CGCL-codes/AdvCLIP}.
\end{abstract}



\maketitle
\section{INTRODUCTION} \label{sec:introduction}
With recent advancements in deep learning, multimodal pre-training has emerged as a promising area of research for various downstream tasks.
Multimodal contrastive learning~\cite{radford2021learning, yuan2021multimodal} 
is a novel machine learning paradigm to overcome the restrictions of labeled data. 
It uses large-scale, noisy, and unprocessed multimodal data pairs sourced from the web to train a cross-modal pre-trained encoder, such as CLIP~\cite{radford2021learning}, with powerful feature extraction capabilities. 
By fine-tuning these pre-trained encoders with a small amount of labeled data, complex and diverse downstream tasks can be performed ~\cite{zeng2022comprehensive, li2023vipmm}. 
This pre-training approach provides a solution for resource-constrained users to benefit from  large-scale models by using their powerful zero-shot capabilities directly or fine-tuning a linear layer for various downstream tasks with less data and computational resources.
Driven by this promising prospect, many service providers have unveiled their pre-trained encoders such as CLIP~\cite{radford2021learning}, ALBEF~\cite{li2021align}, and GPT~\cite{brown2020language}, or have deployed them as commercial services, like ChatGPT.

 \begin{figure}[!t]
    \centering
    \includegraphics[scale=0.49]{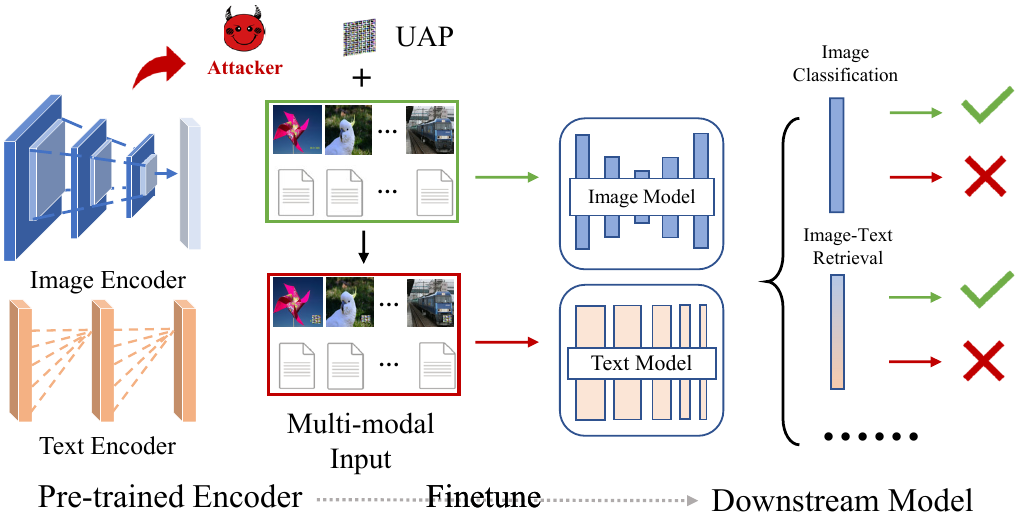}
    \caption{Illustration of attackers using a cross-modal pre-trained encoder to attack different downstream tasks}
    \label{fig:pipeline}
      \vspace{-0.3cm}
\end{figure}

Meanwhile, it is well-known that machine learning models are susceptible to various adversarial attacks~\cite{goodfellow2014explaining,yang2020design,moosavi2017universal}, 
 which will make pre-trained encoders fragile as well.
With pre-trained encoders are widely used, the risks associated with them are often inherited by downstream tasks. 
Recent works~\cite{jia2022badencoder, hu2022badhash, liu2022poisonedencoder, liu2021encodermi, cong2022sslguard, zhou2023advencoder} paid attention to the privacy and robustness concerns of unimodal pre-trained encoders, however, the security threat of more widely used cross-modal pre-trained encoders (\eg, \textit{Vision-language Pre-trained} (VLP) encoders~\cite{radford2021learning, khan2021exploiting}) remains unexplored. 
Although a recent study~\cite{zhang2022towards} tried to conduct adversarial attacks against downstream tasks of VLP encoders, it relied on unrealistic white-box assumptions to generate sample-specific adversarial examples.
In the literature, the difficulty of cross-modal attacks, caused by the heterogeneity between different modalities, has created an illusory sense of security for cross-modal pre-trained encoders.
It tends to become a common belief that it is   impossible  to  realize  cross-modal attacks without the knowledge of the pre-training dataset, the downstream dataset, task type, or even the  defense strategy that the downstream model is taking.
\textit{To the best of our knowledge, implementing adversarial attacks in practical multimodal pre-training scenarios remains a challenging and unsolved problem.}

In this paper, we propose AdvCLIP, the first attack framework for generating downstream-agnostic adversarial examples, to break the existing illusion of security in cross-modal pre-trained encoders. 
Given the limited knowledge of attackers and the feasibility of attack implementation, the goal is to achieve universal non-targeted attacks based on images for downstream tasks.
There are two types of universal adversarial attacks: perturbation-based and patch-based methods.
The former requires adding perturbations to the image globally, the latter is limited to a small area of the image and is more easily applicable to the physical world. Therefore, we mainly focus on adversarial patch attacks. The most daunting challenge in this work is to effectively tackle the modality gap between image and text, while simultaneously bridging the attack gap between cross-modal pre-trained encoders and downstream tasks.

Based on the intuition of maximizing the distance between the target image features and their corresponding benign image and text features, 
we first construct a topological graph structure to capture the similarity between samples.
Then, we fool the pre-trained encoders by destroying the mapping relationship between different modalities of a single sample and the topological relations between multiple samples, respectively.
To achieve attack transferability from the pre-trained encoder to the downstream task, 
we make the adversarial examples far from the original class rather than simply crossing the decision boundary. 
As a result, 
we design a topology-deviation based generative adversarial network to generate a universal adversarial patch to achieve high attack success rate attacks for downstream tasks with a fixed random noise as input.
Our main contributions are summarized as follows:
\begin{itemize}
\item We propose AdvCLIP, the first attack framework to construct downstream-agnostic adversarial examples in multimodal contrastive learning. We reveal that the cross-modal pre-trained encoder incurs severe security risks for the downstream tasks.
\item We design a topology-deviation based generative adversarial network, which adds a universal adversarial patch to the target image,
 to decrease the similarity between different modal embeddings and disrupt their topological relationships, achieving non-targeted adversarial attacks.
\item Our extensive experiments on two types of downstream tasks over eight datasets show that our AdvCLIP addresses the modality gap and transferability between the pre-trained encoders and downstream tasks.
\item We tailor three popular defenses to mitigate AdvCLIP. The
results further prove the attack ability of AdvCLIP and highlight the needs of new defense mechanism to defend pre-trained encoders. 

\end{itemize}
\section{RELATED WORK} \label{sec:related-works}
  
 \subsection{Vision-Language Pre-trained Models}
Multimodal contrastive learning is a training paradigm that aims to pre-train encoders on large-scale unlabeled training data to obtain general-purpose representations for application to downstream tasks.
The success of multimodal contrastive learning has motivated the development of numerous VLP models~\cite{li2020oscar, desai2021virtex, geigle2022retrieve, radford2021learning} for building multimodal models capable of learning vision-language semantic alignments and solving complex cross-modal tasks~\cite{zeng2022comprehensive}.
Existing VLP models can be broadly categorized into two groups: cross-encoder based and embedding-based methods. Cross-encoder based methods~\cite{chen2019uniter, li2020oscar, lu2019vilbert} employ a Transformer-based cross-attention mechanism to compute the similarity between data from different modalities. In contrast, embedding-based methods~\cite{geigle2022retrieve, radford2021learning,sun2021lightningdot} encode data from different modalities separately to generate high-dimensional visual and textual representations and measure cross-modal similarity by computing feature distances between data from different modalities. 
Recently, the embedding-based CLIP~\cite{radford2021learning}
has demonstrated exceptional performance on various downstream
tasks. 
In this paper, we focus on the security of CLIP.

\subsection{Universal Adversarial Attack}
\textit{Universal adversarial perturbation} (UAP) ~\cite{moosavi2017universal} was proposed to deceive a target model by applying a single adversarial noise to all input images.
Universal image-based adversarial attacks come in two forms: perturbations~\cite{moosavi2017universal,deng2020universal,hayes2018learning,mopuri2017fast, hu2022protecting,hu2023pointca,Liu2019Perceptual} and patches~\cite{hu2021advhash,brown2017adversarial, karmon2018lavan,tang2023adversarial}.
Perturbation-based methods fool models by adding visually imperceptible noise globally to the image. In contrast, patch-based methods require precise control over each pixel of the image, resulting in visible adversarial patches that are limited to a small region of the image.
On the other hand,
text-based adversarial attacks require a different approach due to the discrete nature of text. Consequently, universal text-based attacks focus on generating imperceptible triggers and linguistic idioms to create adversarial examples \cite{wallace2019universal, behjati2019universal, song2020universal}.
Unfortunately, the current UAP methods are mostly designed for unimodal classification tasks and are insufficient for attacking cross-modal tasks, let alone when the attacker's knowledge about downstream tasks is limited.
As patch-based image adversarial examples are more applicable to real-world scenarios, this paper focuses on universal adversarial patch.
Additionally, researchers have proposed different defenses against adversarial examples, such as data pre-processing~\cite{carlini2016defensive}, adversarial training~\cite{madry2017towards,tramer2019adversarial,shafahi2020universal}, and pruning~\cite{zhu2017prune,ye2018defending}.

\subsection{Adversarial Attacks on Pre-trained Encoders}
Recently, an increasing number of works~\cite{nern2022adversarial,fan2021does,zhu2022most} begin to investigate the robustness of pre-trained encoders. 
PAP~\cite{ban2022pre} produced a pre-trained perturbation by lifting the feature activations of low-level layers against image pre-trained encoders.
At the same time, some works made trivial attempts to explore the vulnerability of cross-modal pre-trained encoders.
One recent study~\cite{zhu2022most} examined the robustness of a CLIP-based image-text retrieval system.
Furthermore, Co-Attack~\cite{zhang2022towards} took a step by considering the robustness of downstream tasks corresponding to cross-modal pre-trained encoders. It proposed a cooperative loss function to avoid conflicts caused by simultaneous attacks on both image and text modalities. 
However, it only considered simple white-box scenarios where the attacker has downstream knowledge to generate sample-specific adversarial examples.
As a result, insufficient research on cross-modal pre-training safety tends to create a false sense of security in the field.
Our work aims to achieve effective ignorant attacks against downstream tasks and break the illusion of security in cross-modal pre-trained encoders.
\section{METHODOLOGY}\label{sec:method}

\subsection{Threat Model}
We assume a quasi-black-box attack model, where the attacker has access to VLP encoders through purchasing or downloading from publicly available websites, but lacks knowledge of the pre-training datasets and downstream tasks. As the attacker does not possess specific target information of downstream tasks, their objective is to conduct non-targeted adversarial attacks that disable or reduce the accuracy of downstream tasks. 
To achieve this, the attacker leverages the pre-trained encoder to design a downstream-agnostic universal adversarial patch that is applicable to various types of input images from different datasets.
Then the adversarial examples can mislead all the downstream tasks that inherit the victim pre-trained encoder, such as image-text retrieval, image classification, etc. 
We assume that the downstream task undertaker (called user hereinafter) is able to fine-tune the linear layer for their cause.
Given the complexity of CLIP training and its powerful zero-shot performance, we believe that users do not need to fine-tune CLIP directly, as doing so would negate the benefits of choosing it in the first place. We also consider a more stringent scenario, in which users employ common defense mechanisms such as adversarial training to improve the robustness of downstream models.

\subsection{Problem Formulation}
Let $\mathcal{D} =\left \{ (x_{i}, y_{i}) \right \}^{N}_{i=1}$ denote a cross-modal dataset with N instances. Here,  $x_{i} = \left \{ (x_{i}^{v}, x_{i}^{t}) \right \}$, where $x_{i}^{v}$ and $x_{i}^{t}$ represent two data modalities, such as image-text pairs, and they both belong to the same label $y_{i}$. Let $L=\left \{ y_{i} \right \}^{C}_{i=1}$ represent the label dataset from $\mathcal{D}$, where $C$ is the number of labels and $C < N$.
Given an input $x_{i} \in \mathcal{D}_{a}$ to a cross-modal pre-trained encoder $ M_{\theta} (\cdot )$ (\ie, CLIP~\cite{radford2021learning}) which consists of an image encoder $ E_{v} (\cdot )$ and a text encoder $ E_{t} (\cdot )$, that returns an image feature vector $v_{v}$ and  a text feature vector $v_{t}$ respectively, 
where $\theta$ denotes the parameter of the cross-modal pre-trained encoder.
The attacker utilizes a surrogate dataset $\mathcal{D}_{a}$ that is distinct from both the pre-training dataset $\mathcal{D}_{p}$ and the downstream dataset $\mathcal{D}_{d}$ to generate a universal adversarial noise against the pre-trained encoder.
Moreover, the universal adversarial noise $\delta$ should be suffciently small, and modeled through an upper-bound $\epsilon$ on the $l_{p}$-norm.
This problem can be formulated as:
\begin{equation} \label{eq:1}
 M_{\theta}\left (  x_{i} + \delta  \right )  \ne  M_{\theta}\left ( x_{i}  \right ), \quad  s.t.\left \| \delta  \right \| _{p}\le \epsilon
\end{equation}

With the help of the strong feature extraction ability of cross-modal pre-trained encoders, we can just fine-tune a linear layer using output feature vectors of different modalities to achieve complex downstream tasks.
In this paper, we mainly consider cross-modal image-text retrieval  and  unimodal image classification tasks. 
For the cross-modal retrieval task, the cross-modal retrieval head $c_{\theta^{'}}(\cdot )$  completes the image-text retrieval task based on the similarity   between $v_{v}$ and $v_{t}$, where $\theta^{'}$ denotes   the parameter of the retrieval head. The attacker's goal is to implement a non-targeted attack that fools the downstream cross-modal retrieval head $c_{\theta^{'}}(\cdot )$  by applying a universal adversarial noise $\delta$ to the downstream  sample $x \in \mathcal{D}_d$.
Therefore, the attacker's goal can be formalized as:
\begin{equation} \label{eq:3}
c_{\theta^{'} }(E_{v}(x_{i}^{v} + \delta), E_{t}(x_{i}^{t})) \ne  c_{\theta^{'} }(E_{v}(x_{i}^{v}), E_{t}(x_{i}^{t})), \quad s.t.  \left \| \delta  \right \|_{p} \le \epsilon
\end{equation}

Similarly, the objective of attackers against a downstream image classification task can be expressed as:
\begin{equation} \label{eq:2}
f_{\theta^{''} }(E_{v}(x_{i}^{v} + \delta)) \ne  f_{\theta^{''} }(E_{v}(x_{i}^{v})), \quad s.t.  \left \| \delta  \right \|_{ p} \le \epsilon
\end{equation}
where $f_{\theta^{''}}(\cdot )$ is a classifier, $\theta^{''}$ is the parameter of the classifier.

\subsection{Intuition Behind AdvCLIP} \label{sec:intuition}
\begin{figure}[!t]
    \centering
    \includegraphics[scale=0.325]{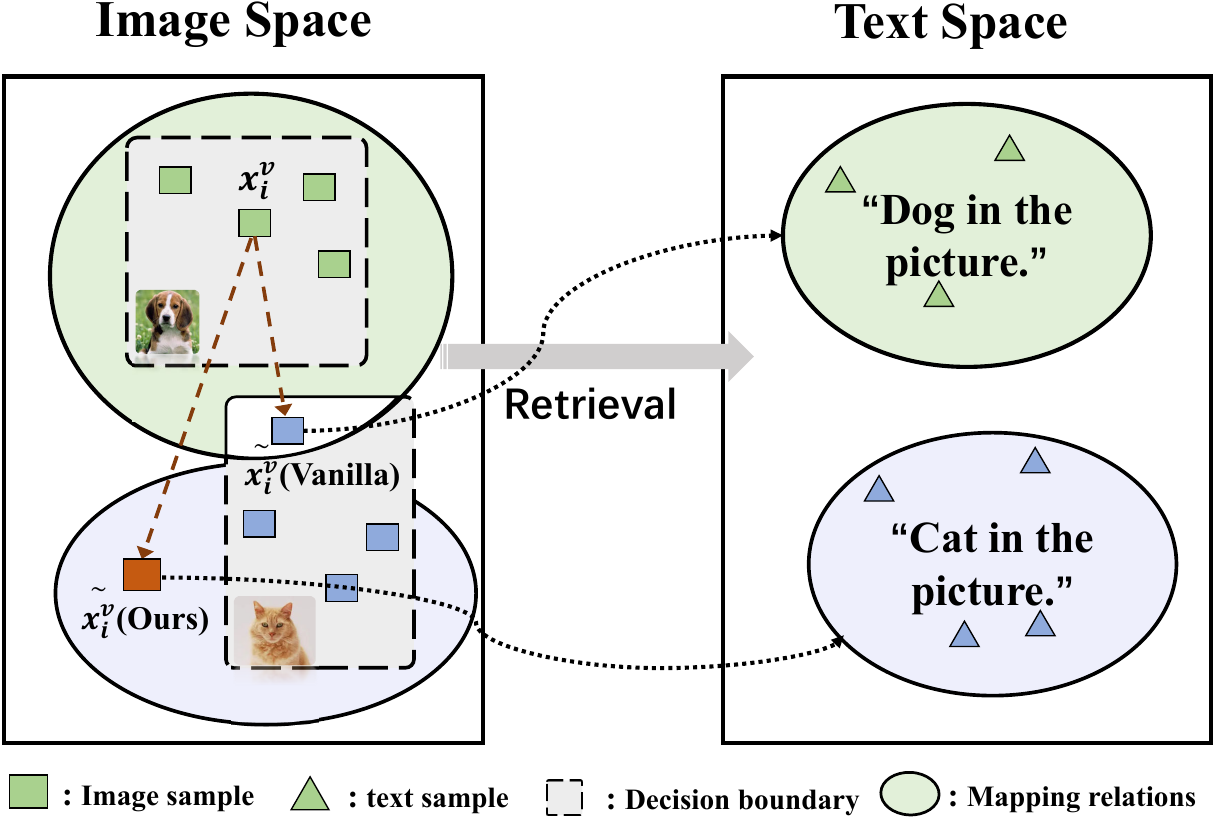}
    \caption{Attack gap between image and text modality}
    \label{fig:cross-gap}
\end{figure}

Due to the limitations of the attacker's knowledge and the complexity of cross-modal tasks, achieving effective attacks on unknown downstream tasks has to address the following challenges:

\textbf{Challenge I: Modality gap between image and text.}
Traditional adversarial attacks are designed for unimodal classification tasks. 
However, the VLP encoder involves multiple modalities and its output is a high-dimensional feature embedding rather than a label, directly applying traditional adversarial attack methods is impractical.
As attackers aim to launch non-targeted adversarial attacks on downstream tasks, a natural idea is to disrupt the similarity matching process by maximizing the feature distance between the adversarial and corresponding clean embeddings of different modalities. 
However, it is a challenging problem to understand and utilize the high-dimensional feature vectors to produce adversarial examples. 
As shown in \cref{fig:cross-gap}, simply maximizing the distance between embeddings (Vanilla) may not work due to the complexity of the high-dimensional feature space and the heterogeneity of multimodal data.
Even if the image adversarial example leaves its original category in the image feature space and is classified as a cat, yet it will still be retrieved to match the textual information of the original category of dogs.

 \begin{figure}[!t]
  \setlength{\belowcaptionskip}{-0.5cm}  
    \centering
    \includegraphics[scale=0.325]{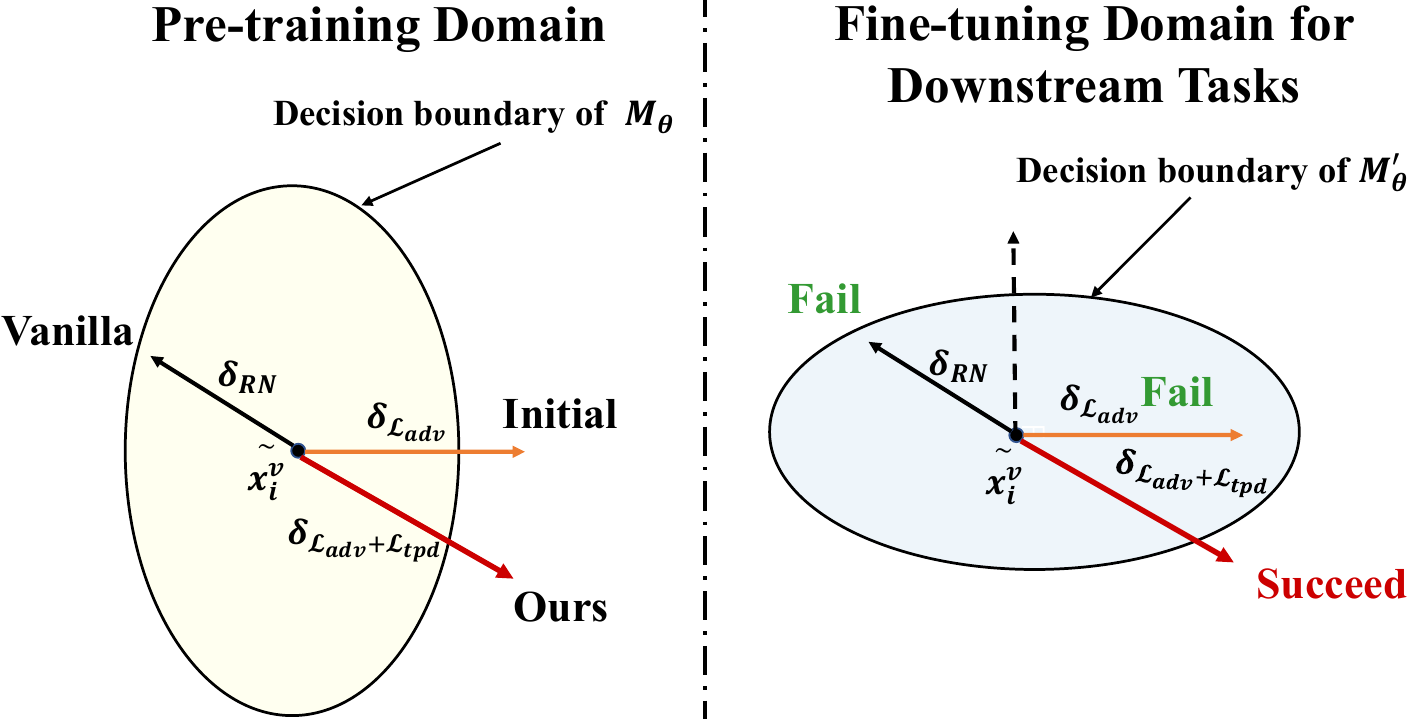}
    \caption{Transferability gap between cross-modal pre-trained encoders and downstream tasks}
    \label{fig:down-gap}
\end{figure}

In this paper, on the basis of leaving the original position in the feature space, we consider destroying the nearest neighbor relationship of the samples to better reinforce the attack by making the ordered samples in the feature space disordered. 
Specifically,
we first construct topology for both adversarial and benign embeddings separately to measure the corresponding sample correlations.
Topology is based on the neighborhood relation graph constructed by the similarity between samples in the representation space. 
The process of measuring topological similarity can be formalized as:
\begin{equation} \label{eq:4}
\mathcal{L}_{tp} = \mathbb{E}_{\left ( x,y \right )\in \mathcal{D}_{a} } \left ( CE\left (\mathcal{G}_{nor}, \mathcal{G}_{adv}  \right )  \right ) 
\end{equation}
where $\mathcal{G}_{nor}$ and $\mathcal{G}_{adv}$ stand for the neighbourhood relation graph constructed by the inter-sample similarity for clean samples and adversarial examples, respectively. $CE(\cdot )$ is the cross-entropy loss to measure the similarity  of two graphs.

We define the edge weights of the neighborhood graph as the probability that two different samples are neighbors, and the deviation of the topological structure is achieved by warping the probability distributions of  two graphs. 
Then, we model the conditional probability distribution using an affinity measure based on cosine similarity to construct the adjacency graph, and remove the nearest neighbor points to prevent isolated subgraphs formed by data points with excessively high local density, thereby ensuring the local connectivity of the manifold and better preserving the global structure. 
The process of constructing the adjacency graph can be represented as:

\begin{equation} \label{eq:4}
\mathcal{G} = \left \{ p_{i  \mid j } \bigg| p_{i\mid j } = \frac{\left (2-\left (  d_{ij} - \rho _{j}\right )   \right ) }{ {\textstyle \sum_{k=1,k\ne j}^{N}}\left ( 2-\left (  d_{jk} - \rho _{j}\right )  \right )  } ,0< i,j \le N \right \} 
\end{equation}
where $ p_{i\mid j }$ is the conditional probability that the $i_{th}$ natural sample is the neighbor of the $j_{th}$ natural sample in the feature space of $\mathcal{G}$, $\rho_{j}$ represents the cosine distance from the  $j_{th}$ data point to its nearest neighbor,  $d_{ij}$ denotes the cosine distance between the corresponding embeddings of the two samples.
By deviating from the two dimensions of the sample itself and the nearest neighbor relationship, we destroy the similarity mapping relationship between the sample and its counterpart to achieve an effective attack.

\textbf{Challenge II: Transferability gap between cross-modal pre-trained encoders and downstream models.}
As illustrated in ~\cref{fig:down-gap}, 
after fine-tuning the cross-modal pre-trained encoder to the downstream model, the boundary of the feature space in the model may change,
which could make existing attacks ineffective. 
Therefore, we aim to deviate adversarial examples from the direction most likely to cross their original category boundaries within the given perturbation budget.
To address this challenge, we are motivated to make adversarial examples deviate from the direction that is most likely to leave their original category boundaries under the same perturbation budget.
Inspired by the fact that generative adversarial networks can generate samples with similar salient features~\cite{goodfellow2020generative}, 
we design a generative adversarial network to generate a universal adversarial noise with strong commonality, such that the adversarial examples are far from the original category rather than only just crossing the decision boundary of that category.
In this way, even if the users fine-tune the pre-trained encoder to the downstream model, the adversarial examples still cannot be recognized properly.

 \begin{figure*}[!h]
    \setlength{\abovecaptionskip}{2pt}
    \centering
    \includegraphics[scale=0.48]{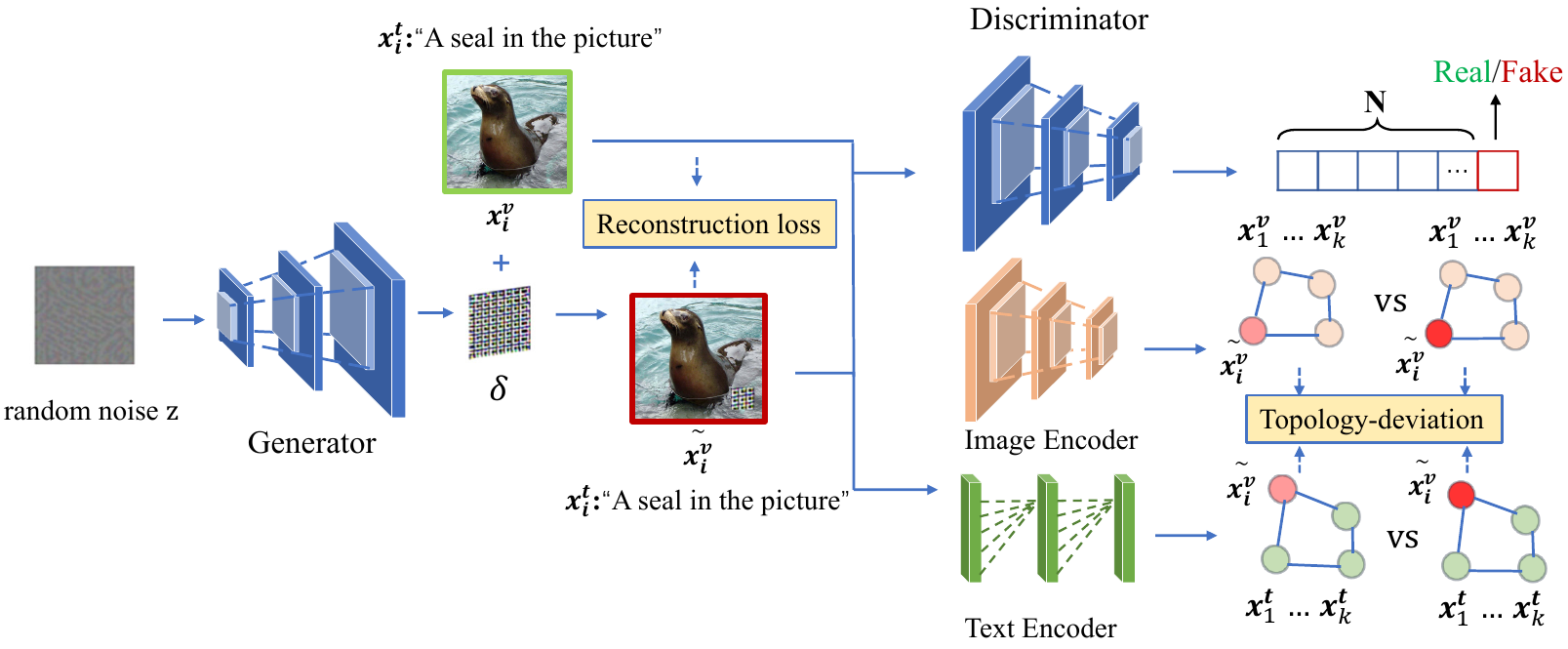}
    \caption{The framework of our attack}
    \label{fig:pipeline}
\end{figure*}

\subsection{Topology-deviation based Generative Attack Framework}
In this section, we present AdvCLIP, a novel generative attack against cross-modal pre-trained encoders. The framework of Adv-CLIP is depicted in ~\cref{fig:pipeline}. It consists of an adversarial generator $G$, a discriminator $D$, and a victim cross-modal encoder $M$ which consists of an image encoder $E_{v}$ and a text encoder $E_{t}$. 
Given the image-text pairs $(x_{i}^{v}, x_{i}^{t})$ to the cross-modal pre-trained encoder, the image encoder $E_{v}$  and text encoder $E_{t}$ output the corresponding feature vectors.
We design a topology-deviation based generative attack framework,
which utilizes cross-modal pre-trained encoders to generate universal adversarial patches applicable to images, thereby deceiving downstream tasks.

\noindent\textbf{Adversarial Generator.} 
By feeding a fixed noise $z$ into the adversarial generator, we obtain a universal adversarial patch $ G(z)$ and paste it onto an image of the surrogate dataset $\mathcal{D}_{a}$ to get an adversarial example $\widetilde{x_{i}^{v}}$. 
The above process of making adversarial examples can be formalized as:
\begin{equation} \label{eq:4}
 \widetilde{x_{i}^{v}}  = x_{i}^{v}  \odot (1- m) + G(z)  \odot m
\end{equation}
where $ \odot$ denotes the element-wise product, $m$ is a binary matrix that contains the position information of the patch.

The objective function of the adversarial generator $G$ is:
\begin{equation} \label{eq:6}
\begin{aligned}
\displaystyle \min_{\theta_{G}}\mathcal{L}_{G} &= \sum_{(x_{i},y_{i})\in \mathcal{D}_{a}}\left ( \alpha \mathcal{L}_{adv} + \beta \mathcal{L}_{tpd} + \mathcal{L}_{q} + \mathcal{L}_{gan}\right )
\end{aligned}
\end{equation}
where  $\mathcal{L}_{adv}$ is the adversarial loss function, $\mathcal{L}_{tpd}$  is the topology-deviation loss function, $\mathcal{L}_{q}$ is the quality loss function, $\mathcal{L}_{gan}$ is the GAN loss function, and $\alpha$, $\beta$ are pre-defined hyper-parameters.

The adversarial loss $\mathcal{L}_{adv}$ is used to deviate the feature position of the target sample, by adding a patch to an image $x_{i}^{v}$ so that the feature vector $E_{v}( \widetilde{x_{i}^{v}})$ of the adversarial example $\widetilde{x_{i}^{v}}$ is simultaneously far away from the original image feature vector $E_{v}(x_{i}^{v})$ and the clean text feature vector $E_{t}(x_{i}^{t})$.
Thus $\mathcal{L}_{adv}$  is expressed as:
\begin{equation}  \label{eq:6}
\mathcal{L}_{adv} = \mathcal{L}_{av} + \lambda \mathcal{L}_{at}
\end{equation}
where $\mathcal{L}_{av}$ and $\mathcal{L}_{at}$ denote the image-image semantic feature deviation loss and the image-text semantic deviation loss, respectively.
We adopt InfoNCE \cite{oord2018representation} loss to measure the similarity between the vectors output by encoders. 
Specifically, we first treat the vector of benign image $x_{i}^{v}$ and  adversarial image $\widetilde{x_{i}^{v}}$ as negative pairs, pulling away their feature distance. It can be expressed as:

\begin{equation}  \label{eq:6}
\mathcal{L}_{av} =  log\left [ \frac{exp\left ( Sim \left (  E_{v}( \widetilde{x_{i}^{v}} ), E_{v}(x_{i}^{v}) \right )  \right /\tau ) }{ {\textstyle \sum_{j=0}^{K}}exp\left (  Sim \left ( E_{v}( \widetilde{x_{i}^{v}} ), E_{v}(x_{j}^{v}) \right / \tau)  \right )  }  \right ]
\end{equation}
where  
$Sim\left ( \cdot \right )$ represents the cosine distance function, 
$\tau$ denotes a temperature parameter.  Then we treat the vectors of adversarial image samples $\widetilde{x_{i}^{v}}$ and benign text samples $x_{i}^{t}$ as negative pairs similarly and increase their feature distances. So we have:

\begin{equation}  \label{eq:6}
\mathcal{L}_{at} =  log\left [ \frac{exp\left ( Sim \left (  E_{v}( \widetilde{x_{i}^{v}} ), E_{t}(x_{i}^{t} ) \right )  \right /\tau ) }{ {\textstyle \sum_{j=0}^{K}}exp\left (  Sim \left ( E_{v}( \widetilde{x_{i}^{v}} ), E_{t}(x_{j}^{t}) \right / \tau)  \right )  }  \right ]
\end{equation}

The topology-deviation loss $\mathcal{L}_{adv}$  is designed to corrupt the topological similarity between the adversarial examples and their corresponding normal samples, \ie, the neighbourhood relation graph constructed based on the similarity between samples in the representation space.
Similarly, we deviate both the image feature vectors $E_{v}( \widetilde{x_{i}^{v}})$ of the adversarial examples and the image feature vectors $E_{v}(x_{i}^{v})$ and text feature vectors $E_{t}(x_{i}^{t})$ of the normal samples. Our goal is to maximize the topological distance between them, which can be represented as:

\begin{equation}  \label{eq:6}
\mathcal{L}_{tpd} = -(\mathcal{L}_{tp}( E_{v}( \widetilde{x_{i}^{v}}),  E_{v}(x_{i}^{v}) )+ \lambda \mathcal{L}_{tp}( E_{v}( \widetilde{x_{i}^{v}}), E_{t}(x_{i}^{t})))
\end{equation}

To achieve better stealthiness, we use $\mathcal{L}_{q}$ to control the magnitude of the adversarial noises output by the generator and crop $\delta$ after each optimisation to ensure it meets the constraints $\varepsilon$. Formally, we have:
\begin{equation} \label{eq:8}
\mathcal{L}_{q} = {\left \|\widetilde{x_{i}^{v}}- x_{i}^{v}  \right \| } _{2}
\end{equation}

The GAN loss $\mathcal{L}_{gan}$ encourages adversarial examples to be more visually natural.
That is, an normal image and an adversarial example with adversarial patch tend to be consistent on the discriminator. 
Thus the  GAN loss $\mathcal{L}_{gan}$ can be expressed as:
\begin{equation}  \label{eq:6}
\mathcal{L}_{gan} =  \log\left (1 - D(\widetilde{x_{i}^{v}}) \right )
\end{equation}

\noindent\textbf{Discriminator.} 
The main function of discriminator is to identify the authenticity of fake examples generated by the adversarial generator.  By playing games with the generator, we ensure that the generated fake adversarial examples are visually indistinguishable from the real ones. The objective loss function of $D$ is:

\begin{equation} \label{eq:10}
\displaystyle \min_{\theta_{D}}\mathcal{L}_\mathbf{{D}} = \sum_{( x_{i},y_{i})\in \mathcal{D}_{a}} -( \log\left (D(x_{i}^{v}) \right )  +  \log\left (1 - D(\widetilde{x_{i}^{v}}) \right ) )
\end{equation}
\section{EXPERIMENTS}

\begin{table*}[htbp]
  \centering
  \caption{The cross-modal attack performance (\%) of AdvCLIP under different settings. $\mathcal{D}_{1}$ - $\mathcal{D}_{4}$ denote the settings where the downstream datasets are NUS-WIDE, Pascal-Sentence, Wikipedia, and XmediaNet, respectively.}
  \scalebox{0.925}{
    \begin{tabular}{ccrrrrrrrrrrrrrrr}
    \toprule[1.5pt]
    \multirow{2}{*}{Surrogate} & \multirow{2}{*}{Dataset} & \multicolumn{3}{c}{ResNet50} & \multicolumn{3}{c}{ResNet101} & \multicolumn{3}{c}{ViT-B/16} & \multicolumn{3}{c}{ViT-B/32} & \multicolumn{3}{c}{ViT-L/14} \\
  \cmidrule(lr){3-5}\cmidrule(lr){6-8} \cmidrule(lr){9-11} \cmidrule(lr){12-14}\cmidrule(lr){15-17}        &       & \multicolumn{1}{c}{$ASR_{i}$} & \multicolumn{1}{c}{$ASR_{t}$} & \multicolumn{1}{c}{AVG} & \multicolumn{1}{c}{$ASR_{i}$} & \multicolumn{1}{c}{$ASR_{t}$} & \multicolumn{1}{c}{AVG} & \multicolumn{1}{c}{$ASR_{i}$} & \multicolumn{1}{c}{$ASR_{t}$} & \multicolumn{1}{c}{AVG} & \multicolumn{1}{c}{$ASR_{i}$} & \multicolumn{1}{c}{$ASR_{t}$} & \multicolumn{1}{c}{AVG} & \multicolumn{1}{c}{$ASR_{i}$} & \multicolumn{1}{c}{$ASR_{t}$} & \multicolumn{1}{c}{AVG} \\
    \hline
    \multirow{4}[1]{*}{NUS-WIDE} & $\mathcal{D}_{1}$    & 45.20  & 17.00  & 31.10  & 36.40  & 4.20  & 20.30  & 67.25  & 57.55  & 62.40  & 43.50  & 8.15  & 25.82  & 45.50  & 5.25  & 25.38  \\
          & $\mathcal{D}_{2}$     & 67.00  & 58.50  & 62.75  & 22.50  & 37.00  & 29.75  & 66.00  & 65.50  & 65.75  & 52.00  & 43.00  & 47.50  & 31.00  & 19.50  & 25.25  \\
          & $\mathcal{D}_{3}$     & 45.24  & 43.73  & 44.48  & 30.09  & 18.40  & 24.25  & 54.76  & 62.34  & 58.55  & 27.49  & 28.57  & 28.03  & 32.25  & 16.45  & 24.35  \\
          & $\mathcal{D}_{4}$     & 57.10  & 47.94  & 52.52  & 59.45  & 35.89  & 47.67  & 80.05  & 63.15  & 71.60  & 65.45  & 42.41  & 53.93  & 54.07  & 11.95  & 33.01  \\
          \hline
    \multirow{4}[0]{*}{Pascal} & $\mathcal{D}_{1}$     & 23.25  & 8.15  & 15.70  & 25.90  & 2.55  & 14.22  & 62.65  & 60.00  & 61.33  & 32.70  & 5.45  & 19.07  & 37.75  & 4.25  & 21.00  \\
          & $\mathcal{D}_{2}$     & 36.00  & 26.00  & 31.00  & 31.00  & 22.00  & 26.50  & 67.50  & 63.00  & 65.25  & 49.50  & 43.50  & 46.50  & 53.50  & 56.00  & 54.75  \\
          & $\mathcal{D}_{3}$    & 13.86  & 17.32  & 15.59  & 13.63  & 9.52  & 11.57  & 51.08  & 55.20  & 53.14  & 26.63  & 18.39  & 22.51  & 37.01  & 21.42  & 29.21  \\
          & $\mathcal{D}_{4}$     & 30.29  & 18.26  & 24.27  & 49.54  & 18.42  & 33.98  & 80.40  & 62.46  & 71.43  & 63.06  & 32.46  & 47.76  & 79.27  & 49.24  & 64.25  \\
          \hline
    \multirow{4}[0]{*}{Wikipedia} & $\mathcal{D}_{1}$     & 32.00  & 7.25  & 19.62  & 33.80  & 0.60  & 17.20  & 63.55  & 53.15  & 58.35  & 23.45  & 13.90  & 18.68  & 51.45  & 38.25  & 44.85  \\
          & $\mathcal{D}_{2}$     & 25.50  & 40.50  & 33.00  & 8.00  & 13.50  & 10.75  & 67.50  & 64.50  & 66.00  & 51.00  & 52.50  & 51.75  & 53.50  & 48.50  & 51.00  \\
          & $\mathcal{D}_{3}$     & 25.11  & 21.00  & 23.05  & 16.45  & 9.96  & 13.21  & 55.84  & 62.99  & 59.41  & 36.15  & 27.27  & 31.71  & 53.46  & 32.03  & 42.74  \\
          & $\mathcal{D}_{4}$     & 34.50  & 29.86  & 32.18  & 47.06  & 20.07  & 33.56  & 80.01  & 63.33  & 71.67  & 56.89  & 33.29  & 45.09  & 82.06  & 56.20  & 69.13  \\
          \hline
    \multirow{4}[1]{*}{XmediaNet} & $\mathcal{D}_{1}$     & 43.20  & 11.30  & 27.25  & 8.66  & 4.33  & 6.50  & 46.53  & 23.16  & 34.84  & 40.05  & 38.52  & 39.28  & 37.66  & 17.96  & 27.81  \\
          & $\mathcal{D}_{2}$     & 59.00  & 62.50  & 60.75  & 36.90  & 7.05  & 21.98  & 58.05  & 5.25  & 31.65  & 45.55  & 3.35  & 24.45  & 33.80  & 9.15  & 21.48  \\
          & $\mathcal{D}_{3}$     & 42.64  & 41.78  & 42.21  & 27.50  & 14.00  & 20.75  & 59.50  & 54.00  & 56.75  & 57.50  & 42.50  & 50.00  & 45.50  & 30.50  & 38.00  \\
          & $\mathcal{D}_{4}$     & 53.76  & 49.03  & 51.40  & 61.66  & 27.46  & 44.56  & 77.84  & 34.90  & 56.37  & 67.32  & 44.06  & 55.69  & 78.58  & 32.73  & 55.66  \\
    \bottomrule[1.5pt]
    \end{tabular}%
    }
  \label{tab:cross_results}%
\end{table*}%

\begin{table*}[htbp]
  \centering
  \caption{The unimodal attack performance (\%) of AdvCLIP under different settings. $\mathcal{V}_{1}$ - $\mathcal{V}_{5}$ denote the settings where the victim models are ResNet50, ResNet101, ViT-B/16, ViT-B/32, and ViT-L/14, respectively.}
    \scalebox{0.9}{
    \begin{tabular}{ccrrrrrrrrrrrrrrrr}
    \toprule[1.5pt]
    \multirow{2}{*}{Surrogate} & \multirow{2}{*}{Victim} & \multicolumn{2}{c}{CIFAR10} & \multicolumn{2}{c}{GTSRB} & \multicolumn{2}{c}{ImageNet} & \multicolumn{2}{c}{NUS-WIDE} & \multicolumn{2}{c}{Pascal} & \multicolumn{2}{c}{STL10} & \multicolumn{2}{c}{Wikipedia} & \multicolumn{2}{c}{XmediaNet} \\
    \cmidrule(lr){3-4} \cmidrule(lr){5-6} \cmidrule(lr){7-8}\cmidrule(lr){9-10}\cmidrule(lr){11-12}\cmidrule(lr){13-14}\cmidrule(lr){15-16}\cmidrule(lr){17-18}             &       & \multicolumn{1}{c}{FR} & \multicolumn{1}{c}{ASR} & \multicolumn{1}{c}{FR} & \multicolumn{1}{c}{ASR} & \multicolumn{1}{c}{FR} & \multicolumn{1}{c}{ASR} & \multicolumn{1}{c}{FR} & \multicolumn{1}{c}{ASR} & \multicolumn{1}{c}{FR} & \multicolumn{1}{c}{ASR} & \multicolumn{1}{c}{FR} & \multicolumn{1}{c}{ASR} & \multicolumn{1}{c}{FR} & \multicolumn{1}{c}{ASR} & \multicolumn{1}{c}{FR} & \multicolumn{1}{c}{ASR} \\
    \hline
    \multirow{6}[2]{*}{NUS-WIDE} & $\mathcal{V}_{1}$    & 89.73  & 65.01  & 90.36  & 67.86  & 94.01  & 75.40  & 76.81  & 56.98  & 78.12  & 50.78  & 71.10  & 63.24  & 81.67  & 45.87  & 91.14  & 78.85  \\
          & $\mathcal{V}_{2}$   & 78.55  & 59.57  & 86.75  & 62.35  & 60.41  & 44.23  & 49.80  & 29.59  & 64.45  & 35.55  & 17.80  & 11.60  & 66.07  & 31.50  & 59.33  & 48.47  \\
          & $\mathcal{V}_{3}$    & 98.00  & 83.49  & 95.85  & 80.30  & 99.73  & 88.80  & 97.37  & 73.06  & 98.50  & 72.00  & 98.04  & 95.56  & 96.00  & 59.68  & 99.96  & 89.78  \\
          & $\mathcal{V}_{4}$    & 87.91  & 73.65  & 91.25  & 68.79  & 96.21  & 77.52  & 68.51  & 49.02  & 91.41  & 61.33  & 78.85  & 70.83  & 77.96  & 41.27  & 91.53  & 79.24  \\
          & $\mathcal{V}_{5}$    & 43.95  & 34.73  & 90.77  & 71.06  & 89.87  & 74.51  & 55.76  & 36.91  & 63.67  & 34.38  & 52.71  & 48.45  & 81.72  & 37.47  & 80.84  & 70.24  \\
          & AVG   & 79.63  & 63.29  & 91.00  & 70.07  & 88.05  & 72.09  & 69.65  & 49.11  & 79.23  & 50.81  & 63.70  & 57.94  & 80.68  & 43.16  & 84.56  & 73.32  \\
    \hline
    \multirow{6}[2]{*}{XmediaNet} & $\mathcal{V}_{1}$     & 86.37  & 61.87  & 96.74  & 74.67  & 87.68  & 72.25  & 75.44  & 52.73  & 80.08  & 50.78  & 71.80  & 68.90  & 65.54  & 34.15  & 90.39  & 80.00  \\
          & $\mathcal{V}_{2}$     & 90.33  & 67.39  & 73.25  & 63.76  & 55.67  & 49.50  & 80.91  & 56.01  & 86.72  & 62.11  & 37.01  & 35.98  & 60.24  & 30.92  & 62.05  & 56.88  \\
          & $\mathcal{V}_{3}$     & 55.03  & 41.34  & 74.89  & 65.51  & 74.08  & 67.39  & 83.20  & 57.67  & 50.00  & 30.08  & 21.76  & 20.93  & 73.49  & 38.48  & 89.74  & 83.96  \\
          & $\mathcal{V}_{4}$     & 83.82  & 69.98  & 91.69  & 81.98  & 95.41  & 88.62  & 87.65  & 62.40  & 89.45  & 59.77  & 78.45  & 77.54  & 90.09  & 57.14  & 98.30  & 92.09  \\
          & $\mathcal{V}_{5}$     & 60.10  & 52.48  & 91.22  & 81.76  & 91.57  & 84.66  & 77.64  & 53.13  & 79.30  & 51.56  & 49.91  & 49.05  & 70.03  & 37.67  & 90.02  & 83.98  \\
          & AVG   & 75.13  & 58.61  & 85.56  & 73.54  & 80.88  & 72.48  & 80.97  & 56.39  & 77.11  & 50.86  & 51.79  & 50.48  & 71.88  & 39.67  & 86.10  & 79.38  \\
    \bottomrule[1.5pt]
    \end{tabular}%
    }
  \label{tab:solo_results}%
\end{table*}%

\subsection{Experimental Setting}

\noindent\textbf{Victim Pre-trained Encoders.} 
We choose CLIP~\cite{radford2021learning} as the victim encoder  for our experiments and obtain all pre-trained encoders from its publicly available repository.
We evaluate the vulnerability of CLIP to adversarial attacks across a range of architectures, including ResNet50, ResNet101, ViT-L/14, ViT-B/16, and ViT-B/32. 

\noindent\textbf{Downstream Datasets.} 
We evaluate the effectiveness of our attacks on two distinct downstream tasks: image-text retrieval and image classification. 
To carry out the image-text retrieval task, we select four widely used cross-modal datasets, namely 
Wikipedia~\cite{rashtchian2010collecting}, Pascal-Sentence~\cite{rasiwasia2010new}, NUS-WIDE~\cite{chua2009nus}, and  XmediaNet~\cite{peng2018modality}. For the image classification task, we additionally choose STL10~\cite{coates2011analysis}, GTSRB~\cite{stallkamp2012man}, CIFAR10~\cite{krizhevsky2009learning}, and ImageNet~\cite{russakovsky2015imagenet} image datasets. Note that our approach is to generate image adversarial patches using the cross-modal datasets.

\noindent\textbf{Evaluation Metrics.}
In the cross-modal retrieval task,
We use the standard evaluation metric, \emph{mean average precision} (MAP)~\cite{zuva2012evaluation}, to evaluate the accuracy of models, which we report separately for two sub-tasks: text retrieval with image queries (I2T) and  image retrieval with text queries (T2I). 
To measure the performance of our attacks, following~\cite{wang2022cross}, we use the \textit{attack success rate} (ASR), which is calculated as the difference between the MAP values of normal samples and adversarial examples, with $ASR_i$ and $ASR_t$ used for image-text retrieval and text-image retrieval, respectively.
For the classification task, we evaluate our attacks using three metrics: \textit{clean accuracy} (CA), \textit{attack success rate} (ASR), and  \textit{fooling rate} (FR). 
CA measures normal accuracy, ASR is calculated as described above, and FR is the percentage of misclassified examples compared to the total number of test examples.
Higher ASR and FR values indicate better attack effectiveness. 

\subsection{Attack Performance} \label{sec:attack-performance}
\begin{figure}[!t]
  \setlength{\abovecaptionskip}{2pt}
    \centering
    \includegraphics[scale=0.3]{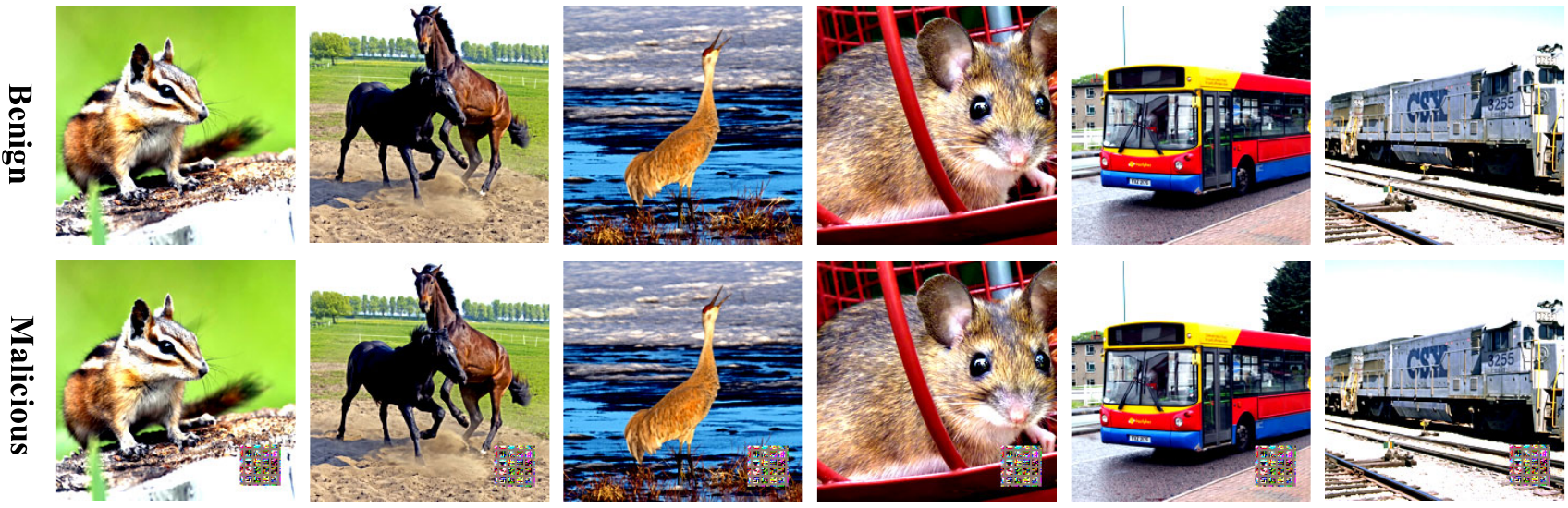}
    \caption{Adversarial examples generated by AdvCLIP based on XmediaNet}
    \label{fig:demo}
    \vspace{-4mm}
\end{figure}

\noindent\textbf{Implementation Details.} 
In order to evaluate the effectiveness of AdvCLIP in scenarios where the downstream task is unknown, we conduct experiments on two distinct tasks: image-text retrieval and image classification. 
Following~\cite{hu2021advhash,moosavi2017universal}, we set the perturbation upper bound (the noise percentage of each sample) $\epsilon$ of adversarial patch to $0.03$. We choose the bottom right corner of the image, which is not easily visible, to apply the patch.
We set the hyper-parameters $\alpha= 10$, $\beta = 5$, $\lambda = 1$ and the training epoch to $20$ with batch size of $16$.
We use four cross-modal datasets as attacker surrogate datasets to train  generative adversarial networks to generate a universal adversarial patch, which are then used to attack different downstream tasks.
The generator and discriminator network are trained by Adam optimizer with the initial learning rate $0.0002$. 
Examples of generated adversarial patches are shown in ~\cref{fig:demo}.

For the image-text retrieval task, we comprehensively evaluate the attack performance of AdvCLIP on four downstream cross-modal datasets (NUS-WIDE, Pascal-Sentence, Wikipedia, and XmediaNe). 
We evaluate the performance of AdvCLIP in two subtasks of image-text retrieval using $ASR_{i}$ and $ASR_{t}$, respectively.
For the image classification task, in addition to using images from the above cross-modal dataset, we also use four commonly used image  datasets (CIFAR10, STL10, GTSRB, and ImageNet) to train the classification model.
We use $ASR$ and $FR$ to evaluate AdvCLIP's ability in the classification task. 

\begin{figure*}[!t]   
  \centering
      \subcaptionbox{Loss-Pascal}{\includegraphics[width=0.19\textwidth]{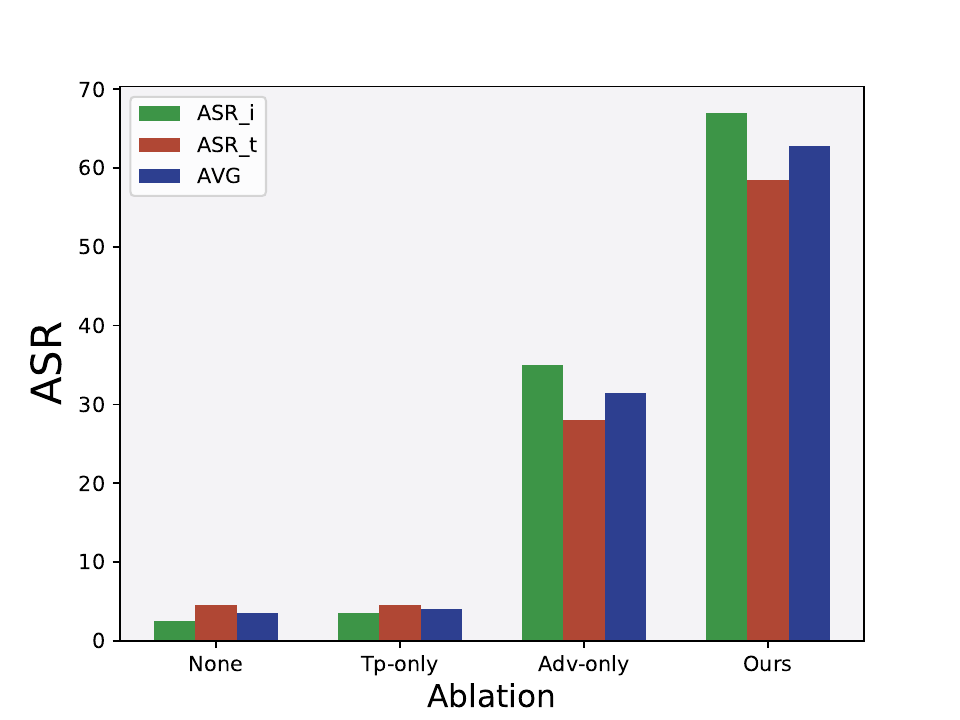}}
     \subcaptionbox{Loss-Wikipedia}{\includegraphics[width=0.19\textwidth]{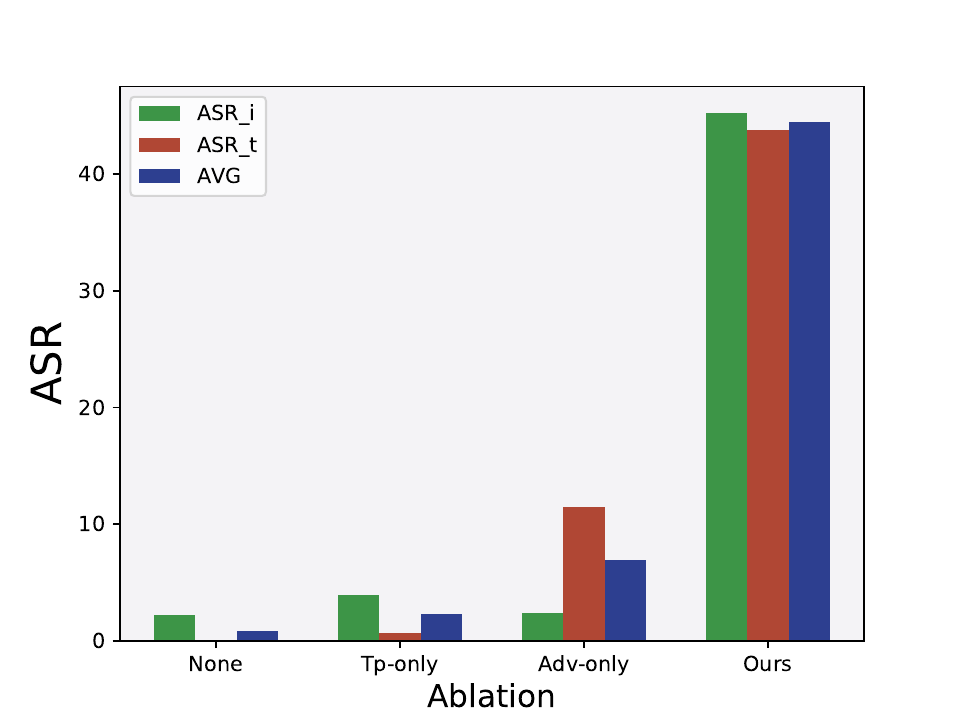}}
             \subcaptionbox{Strength-Pascal}{\includegraphics[width=0.19\textwidth]{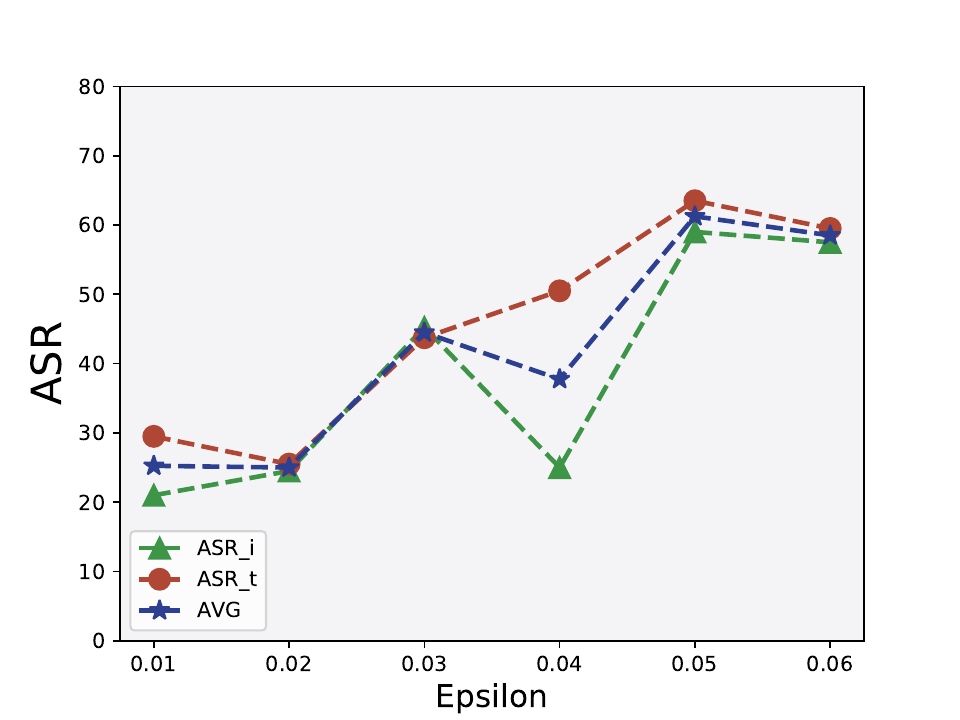}}
     \subcaptionbox{Strength-Wikipedia}{\includegraphics[width=0.19\textwidth]{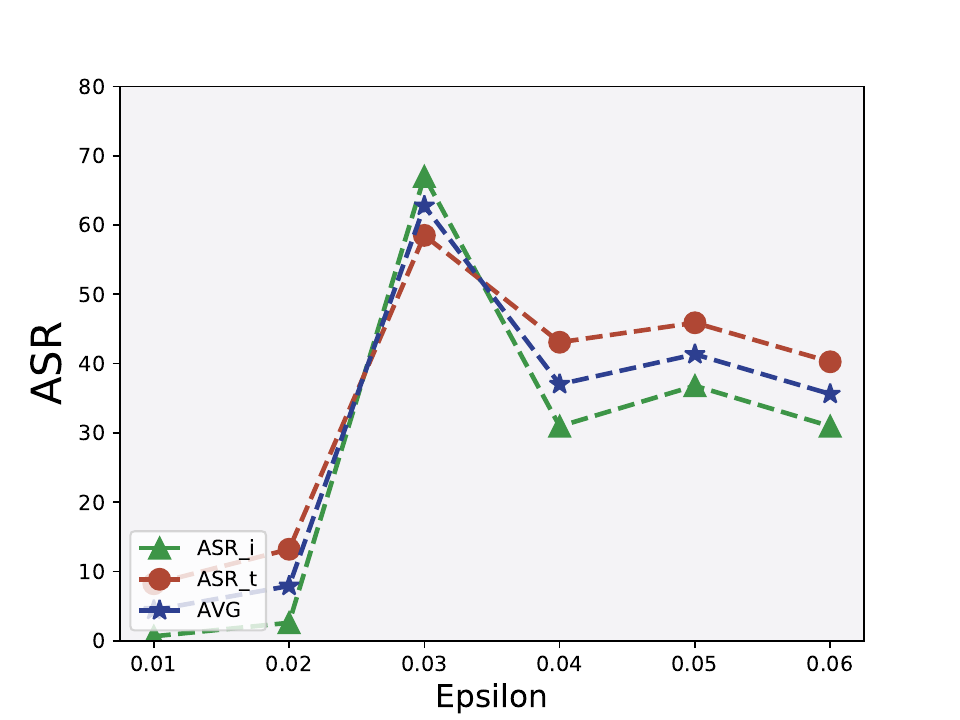}}
       \subcaptionbox{Batch-Pascal}{\includegraphics[width=0.19\textwidth]{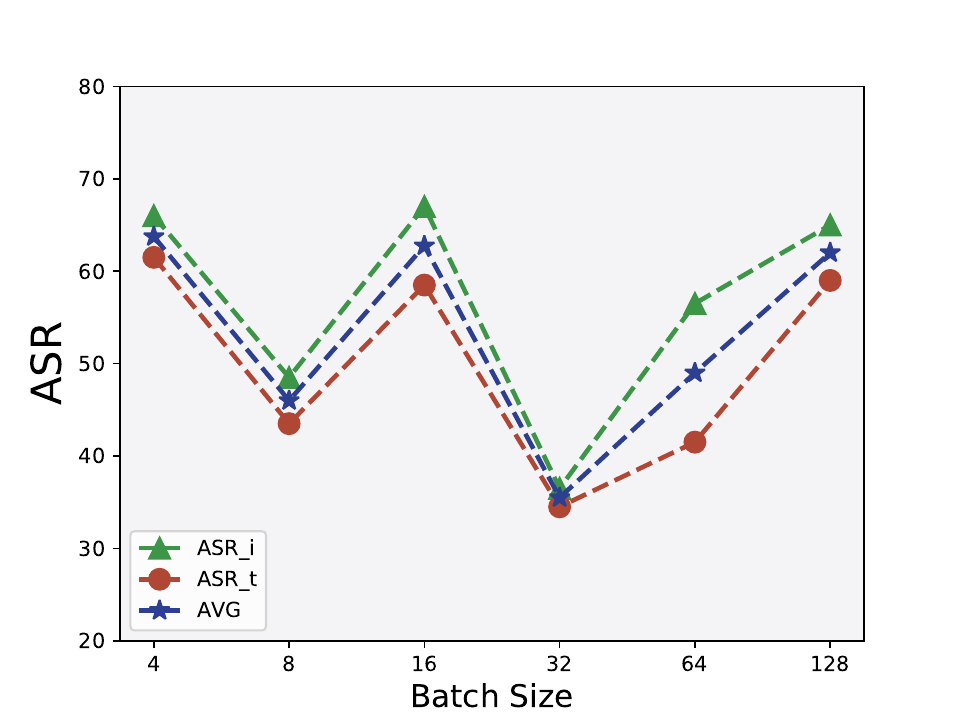}}
     \subcaptionbox{Batch-Wikipedia}{\includegraphics[width=0.19\textwidth]{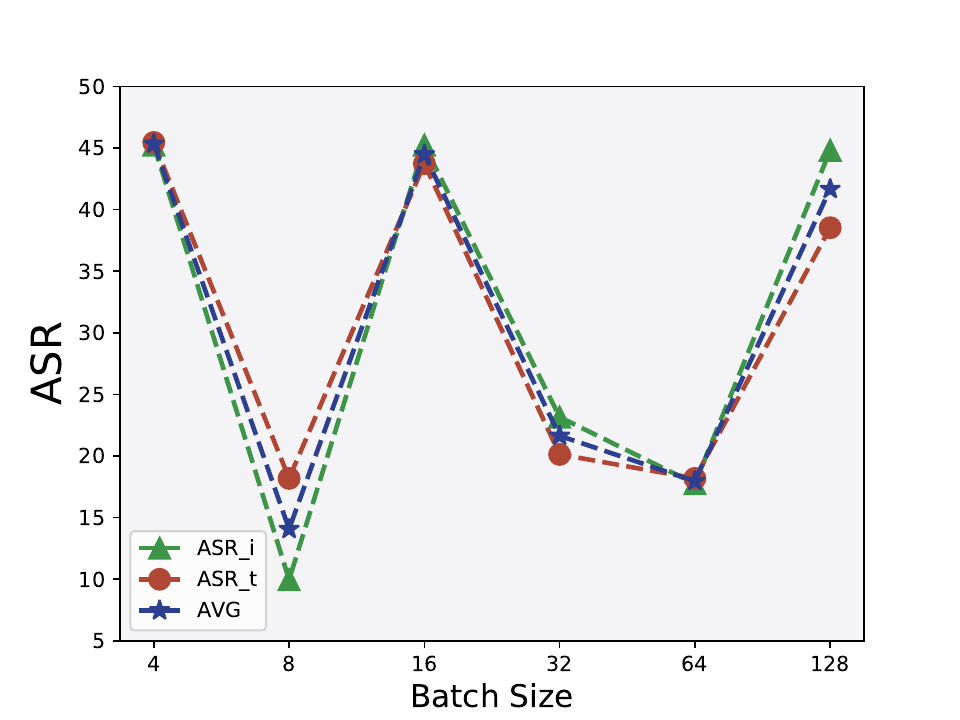}}
        \subcaptionbox{Corruption-Wikipedia}{\includegraphics[width=0.19\textwidth]{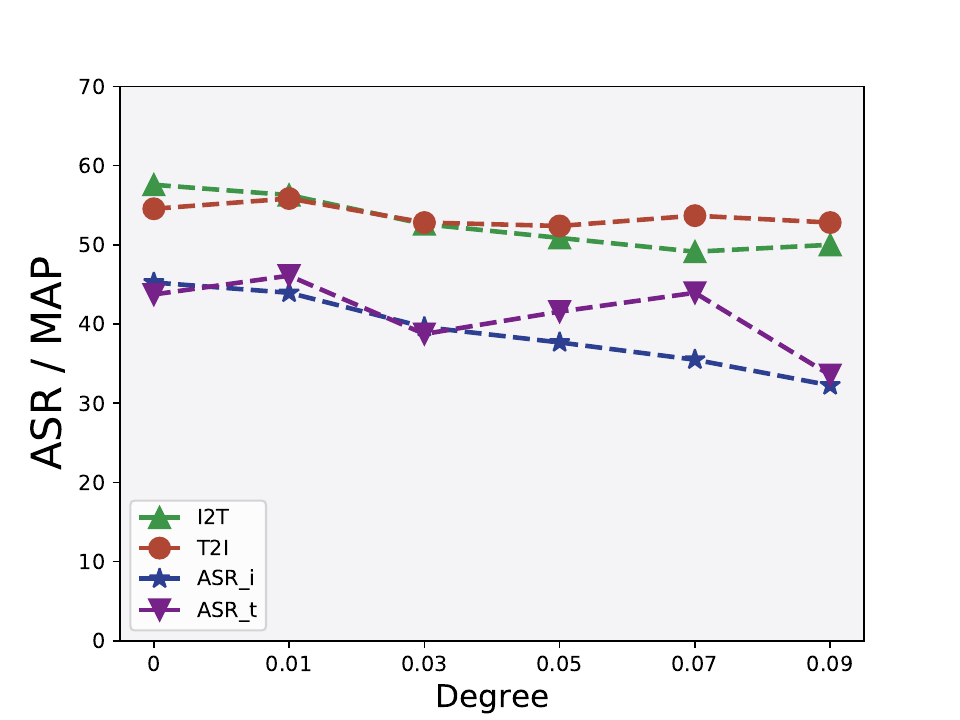}}
     \subcaptionbox{Corruption-Pascel}{\includegraphics[width=0.19\textwidth]{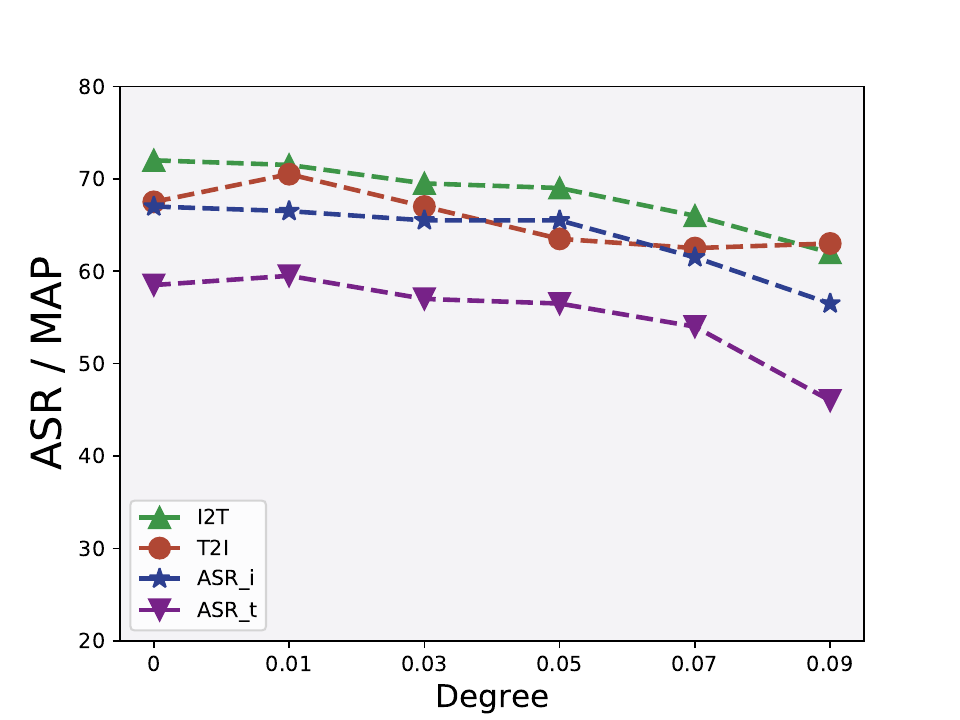}}
           \subcaptionbox{Pruning-Wikipedia}{\includegraphics[width=0.19\textwidth]{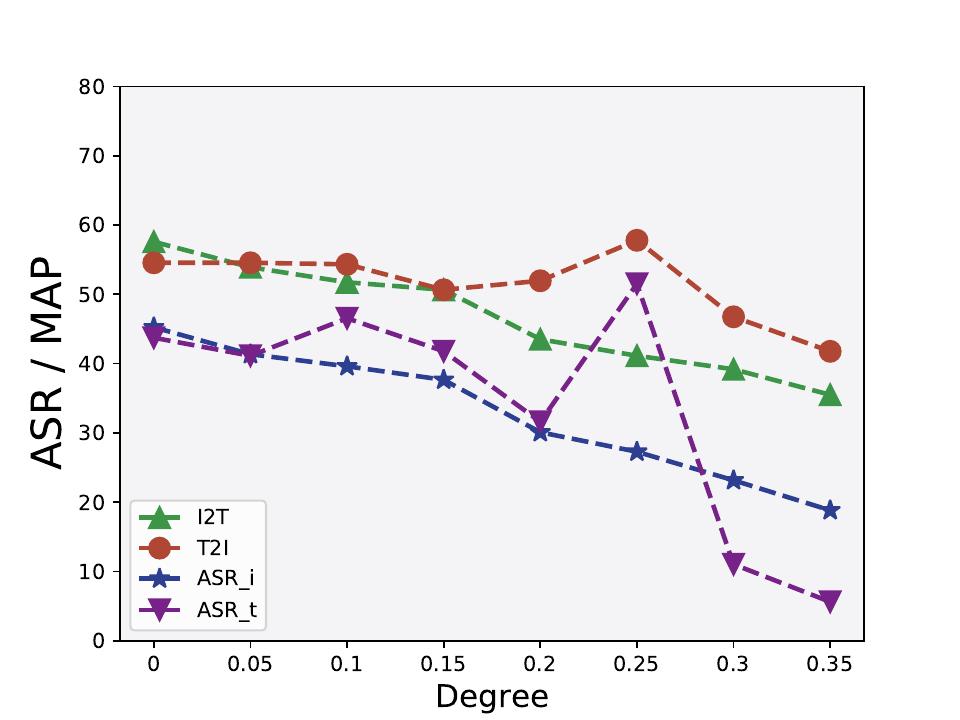}}
     \subcaptionbox{Pruning-Pascel}{\includegraphics[width=0.19\textwidth]{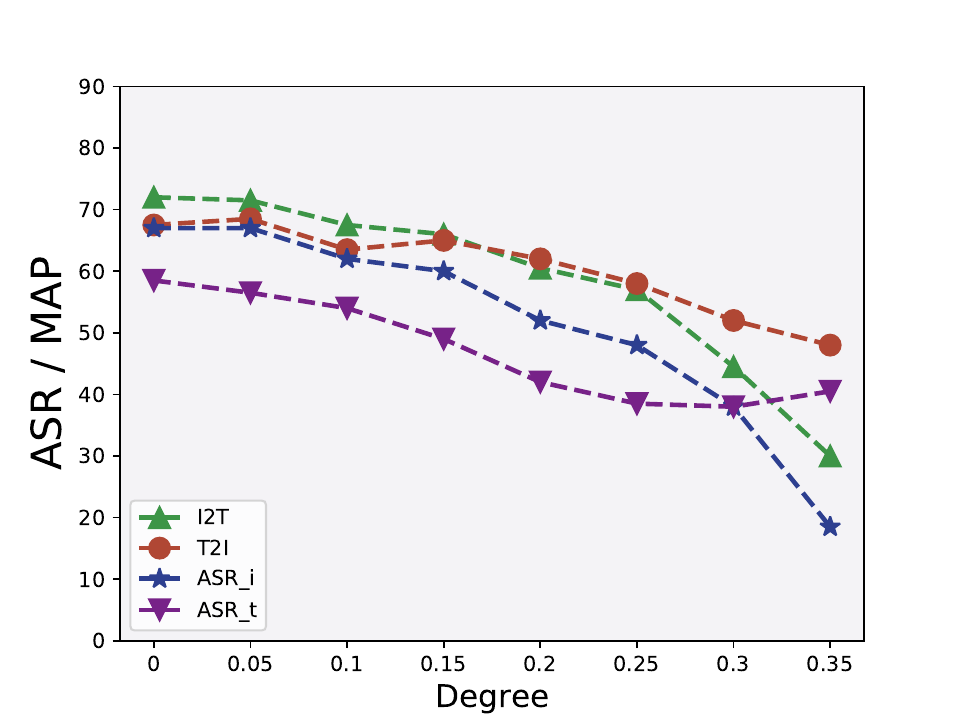}}
      \caption{The attack performance under different settings. (a) - (f) examine the effects of different modules, attack strengths, and batch sizes on AdvCLIP, respectively.
     (g) - (j) investigate the effect of defense methods on AdvCLIP, respectively.}
       \label{fig:ablation results}
\end{figure*}

\noindent\textbf{Analysis.} 
Our results in the image-text retrieval downstream task demonstrate a significant security threat posed by cross-modal pre-trained encoders. 
Firstly, ~\cref{tab:cross_results} shows that different types of backbones have varying vulnerability to adversarial patches under the same attack settings, with the Transformer architecture being more susceptible to successful attacks than ResNet.
Secondly, the surrogate dataset has a significant effect on downstream attack success.
Datasets such as NUS-WIDE and XmediaNet, which contain a larger number of samples, tend to result in higher attack success rates.
Thirdly, the attack performance may not be optimal when the surrogate dataset is consistent with the downstream dataset.
For attackers, creating a better surrogate dataset is an essential factor in achieving success in unknown downstream tasks.
As shown in ~\cref{tab:solo_results}, we achieve impressive performance in the classic image classification task. The average FR value of the output results of the downstream classification model is as high as 70\%, and the average accuracy drop of the model is also over 55\%. 

\subsection{Ablation Study}
In this section, we explore the effect of different modules, surrogate datasets, attack strengths, and batch sizes on AdvCLIP.
For our experiments, we select CLIP based on ResNet50 as the victim encoder and use the NUS-WIDE dataset as the attacker surrogate dataset to launch attacks on image-text retrieval tasks. 

\noindent\textbf{The Effect of $\mathcal{L}_{adv}$ \&  $\mathcal{L}_{tpd}$.} 
We first analyze the effect of $\mathcal{L}_{adv}$  and  $\mathcal{L}_{tpd}$ on our scheme, respectively. 
As shown in~\cref{fig:ablation results} (a-b), ``None'' indicates the removal of both, while ``Adv\_only'' and ``Tp\_only'' indicate the removal of $\mathcal{L}_{adv}$ and $\mathcal{L}_{tpd}$, respectively.  
Our results show that using $\mathcal{L}_{adv}$ alone is not enough to push away the feature position of the sample itself, and the ability to attack the downstream task is poor. However, by simultaneously disrupting the sample's relative position in the feature space, ideal results can be achieved, as indicated by the "Our" results.


\noindent\textbf{The Effect of $\epsilon$.} 
We study the effect of different perturbation upper bound on the attack performance of AdvCLIP. 
From~\cref{fig:ablation results}(c-d), we can see that CLIP has different sensitivities to different perturbation budgets.
A higher attack success rate can be achieved with a smaller patch size when $\epsilon$ is 0.03.

\noindent\textbf{The Effect of Batch Size.} 
We examine the effect of different batch sizes from $4$ to $128$ on AdvCLIP, the results are shown in~\cref{fig:ablation results}(e-f). 
To balance both attack performance and computational efficiency, we set the batch size to 16 as the default setting.

\subsection{Comparison Study}
\noindent\textbf{Implementation Details.} 
In this section, we compare AdvCLIP with \textit{state-of-the-art} (SOTA) universal adversarial attacks. 
Prior researches have not focused enough on downstream tasks for VLP encoders.
The work most relevant to ours is Co-Attack~\cite{zhang2022towards},
which generates sample-specific adversarial examples in a white-box setting. 
To facilitate comparison, 
we randomly select an adversarial perturbation generated by Co-Attack for the images for testing.
Furthermore, to demonstrate the superiority of our approach, we also consider adversarial attacks against unimodal image encoders, including PAP~\cite{ban2022pre} against image pre-trained models and classic universal adversarial perturbation schemes (\eg, UAP~\cite{moosavi2017universal}, UPGD~\cite{deng2020universal}, FFF~\cite{mopuri2017fast}, SSP~\cite{naseer2020self}, and Adv-Patch~\cite{brown2017adversarial}).
To conduct a comprehensive comparison with SOTA schemes under the paradigm of cross-modal pre-trained encoders to downstream tasks, we select two representative architectures, ResNet50 and ViT-B/16, of CLIP and evaluate them on image-text retrieval downstream tasks.

\noindent\textbf{Analysis.}  From \cref{tab:compare_results_cross}, we can see that AdvCLIP outperforms existing SOTA methods by a large margin on two downstream datasets.
Note that Co-Attack needs to use [CLS] in Transformer in the optimization process, so it cannot be used directly to attack CLIP based on ResNet50 (``-").
The negative experimental values (``*") indicate that the attack does not work at all.
There are three reasons for this: 
firstly, CLIP's robustness originates from its pre-training on a vast dataset of 400 million image-text pairs. Secondly, our quasi-black-box threat model limits the attacker's knowledge, making it difficult to attack CLIP's downstream model. As observed in  \cref{tab:compare_results_cross}, existing attacks hardly affect CLIP-based models. 
Lastly, unsuccessful perturbations unintentionally align input samples with CLIP's training set, improving overall accuracy and resulting in negative ASR values for the attack.

\begin{table}[t!]
  \centering
  \caption{Attack performance (\%) of comparison study of downstream cross-modal attacks}
   \scalebox{0.78}{
    \begin{tabular}{ccccccccc}
    \toprule[1.5pt]
    \multirow{3}{*}{Method} & \multicolumn{4}{c}{ResNet50} & \multicolumn{4}{c}{ViT-B/16} \\
\cmidrule(lr){2-5} \cmidrule(lr){6-9}         & \multicolumn{2}{c}{Pascal} & \multicolumn{2}{c}{Wikipedia} & \multicolumn{2}{c}{Pascal} & \multicolumn{2}{c}{Wikipedia} \\
\cmidrule(lr){2-3} \cmidrule(lr){4-5} \cmidrule(lr){6-7}\cmidrule(lr){8-9}         & $ASR_{i}$ & $ASR_{t}$ & $ASR_{i}$ & $ASR_{t}$ & $ASR_{i}$ & $ASR_{t}$ & $ASR_{i}$ & $ASR_{t}$ \\
    \hline
    UAP~\cite{moosavi2017universal}   & 8.00  & 5.00  & 4.12  & 0.22  & *  & 20.00  & 0.21  & 12.56  \\
    UPGD~\cite{deng2020universal} & 2.00  & *  & 1.52  & 2.39  & 1.00  & 4.00  & *  & 6.50  \\
    FFF~\cite{mopuri2017fast}    & 2.00  & 2.00  & 2.39  & 1.30  & *  & 2.00  & 0.21  & 6.28  \\
    SSP~\cite{naseer2020self}   & 2.50  & 3.00  & 8.23  & 4.98  & 4.50  & 2.00  & *  & 6.50 \\
    PAP-ugs~\cite{ban2022pre} & 1.00  & 0.50  & 5.63  & 0.87  & 2.50  & *  & 1.08  & 5.85  \\
    Co-Attack~\cite{zhang2022towards} & -  & -  & -  & -  & 14.50  & 5.00  & 9.30  & 8.23  \\
    Adv-Patch~\cite{brown2017adversarial} & 30.00  & 20.50  & 9.10  & 10.61  & 15.00  & 16.00  & 27.05  & 19.70  \\
    Ours   & \textbf{67.00}  & \textbf{58.50}  & \textbf{45.24}  & \textbf{43.73}  & \textbf{66.00}  & \textbf{65.50}  & \textbf{54.76}  & \textbf{62.34}  \\
    \bottomrule[1.5pt]
    \end{tabular}%
    }
  \label{tab:compare_results_cross}%
\vspace{-4mm}
\end{table}%

\section{Defense}
In this section, we  tailor three downstream defenses for adaptively mitigating AdvCLIP. 
For users utilizing cross-modal pre-trained encoders, they can preprocess input data, conduct adversarial training on downstream models, or prune parameters to defend against adversarial attacks, while mataining normal model accuracy.

\subsection{Corruption}
Corruption is an effective and simple countermeasure for purifying adversarial examples at the pre-processing phase~\cite{carlini2016defensive}. 
To combat adversarial examples, we introduce different levels of Gaussian noise to corrupt input images. 
As shown in ~\cref{fig:ablation results}(g-h),  while mataining that the clean samples maintain normal accuracy, the retrieval accuracy of the model decreases significantly with the degree of corruption, and the performance of AdvCLIP is slightly affected. 
These results indicate that AdvCLIP can effectively resist the corruption-based pre-processing defense.

\subsection{Pruning}
Pruning~\cite{zhu2017prune} is widely used for downstream models to inherit pre-trained encoders by removing redundant parameters in neural networks, reducing model size and computational complexity. 
While pruning the parameters, the required dependencies of the adversarial examples designed for the pre-trained encoder structure and parameters are broken to effectively defend against the adversarial attack.
We perform parameter pruning on CLIP based on ResNet50 and evaluate the effectiveness of our attack using NUS-WIDE as surrogate dataset on two  downstream cross-modal datasets.  The results  in  ~\cref{fig:ablation results}(i-j) show that pruning parameters is difficult to effectively resist CLIP while mataining normal model accuracy.

\subsection{Adversarial Training}
Adversarial training~\cite{goodfellow2014explaining} commonly mixes adversarial examples with original data to enhance model robustness and generalization, making it more resistant to adversarial attacks. We consider a more stringent scenario where defenders conduct adversarial training on downstream tasks. 
Following~\cite{goodfellow2014explaining}, we enhance the robustness of the model during training of downstream tasks by adding noise to the samples and incorporating them in the training process.
For the experiment,
we use NUS-WIDE as the surrogate dataset for evaluation on four downstream datasets.
As shown in~\cref{tab:adversarial_training_cross}, our method is still able to successfully attack downstream tasks that have been enhanced by adversarial training.

\begin{table}[t!]
  \centering
  \caption{Attack performance (\%) on models that have undergone adversarial training}
  \scalebox{0.825}{
    \begin{tabular}{ccccccccc}
    \toprule[1.5pt]
    \multirow{2}[2]{*}{Dataset}  & \multicolumn{2}{c}{RN50} & \multicolumn{2}{c}{RN101} & \multicolumn{2}{c}{ViT-B/16} & \multicolumn{2}{c}{ViT-B/32} \\ \cmidrule(lr){2-3} \cmidrule(lr){4-5}  \cmidrule(lr){6-7} \cmidrule(lr){8-9}      
          & $ASR_{i}$ & $ASR_{t}$ & $ASR_{i}$ & $ASR_{t}$ & $ASR_{i}$ & $ASR_{t}$ & $ASR_{i}$ & $ASR_{t}$ \\
    \hline
    NUS-WIDE & 46.75  & 14.95  & 31.75  & 10.70  & 60.65  & 56.30  & 37.90  & 4.20  \\
    Pascal & 54.50  & 60.00  & 25.00  & 44.50  & 77.00  & 75.50  & 66.50  & 40.00  \\
    Wikipedia & 40.91  & 37.45  & 32.04  & 25.33  & 58.44  & 64.93  & 47.84  & 16.45  \\
    XmediaNet & 69.02  & 46.37  & 63.15  & 35.16  & 79.84  & 63.24  & 71.88  & 40.33  \\
    \bottomrule[1.5pt]
    \end{tabular}%
    }
  \label{tab:adversarial_training_cross}%
\vspace{-0.4cm}
\end{table}%

\section{conclusion}\label{conclusion}
In this paper, we propose the first attack framework to construct downstream-agnostic adversarial examples based cross-modal pre-trained encoders in multimodal contrastive learning. 
We design a topology-deviation based generative adversarial network that generates a universal adversarial patch to fool downstream tasks under strict constraints on attacker's knowledge.
We verify the excellent attack performance of AdvCLIP on two types of downstream tasks over five backbones of CLIP on eight datasets.
We tailor three popular defenses to mitigate AdvCLIP. The results further prove the attack ability of AdvCLIP and highlight the needs of new defense mechanism to defend pre-trained encoders.

\section*{Acknowledgments} Shengshan's work is supported in part by the National Natural Science Foundation of China (Grant No.U20A20177) and Hubei Province Key R\&D Technology Special Innovation Project under Grant No.2021BAA032. 
Minghui's work is supported in part by the National Natural Science Foundation of China (Grant No.62202186)
Shengshan Hu is the corresponding author.

\bibliographystyle{ACM-Reference-Format}
\balance
\footnotesize
\bibliography{acmart}

\end{document}